\documentclass[letterpaper]{article} 
\usepackage{aaai23}  
\usepackage{times}  
\usepackage{helvet}  
\usepackage{courier}  
\usepackage[hyphens]{url}  
\usepackage{graphicx} 
\usepackage{natbib}  
\usepackage{caption} 
\usepackage{algorithm}
\usepackage{algorithmic}

\usepackage{newfloat}
\usepackage{listings}
\DeclareCaptionStyle{ruled}{labelfont=normalfont,labelsep=colon,strut=off} 
\floatstyle{ruled}
\newfloat{listing}{tb}{lst}{}
\floatname{listing}{Listing}
\usepackage[switch]{lineno} 
\usepackage{amssymb}
\usepackage{amsmath}
\usepackage{multirow}
\usepackage{booktabs}
\newcommand{\etal}{\textit{et al}. }
\newcommand{\Tref}[1]{Table~\ref{#1}}
\newcommand{\Eref}[1]{Eq.~(\ref{#1})}
\newcommand{\Fref}[1]{Fig.~\ref{#1}}

\title{Pseudo Label-Guided Model Inversion Attack via Conditional Generative Adversarial Network}
\author{
    Xiaojian Yuan\textsuperscript{\rm 1},
    Kejiang Chen\footnote{Corresponding authors.}\textsuperscript{\rm 1},
    Jie Zhang\textsuperscript{\rm 1,2},
    Weiming Zhang\textsuperscript{\rm 1}, \\
    Nenghai Yu\textsuperscript{\rm 1},
    Yang Zhang\textsuperscript{\rm 3}
}
\affiliations{
    
    \textsuperscript{\rm 1}University of Science and Technology of China,
    \textsuperscript{\rm 2}University of Waterloo,\\
    \textsuperscript{\rm 3}CISPA Helmholtz Center for Information Security \\

    xjyuan@mail.ustc.edu.cn, chenkj@ustc.edu.cn, jiezhangsp@gmail.com, \\ zhangwm@ustc.edu.cn, 
    ynh@ustc.edu.cn, zhang@cispa.de

}

\usepackage{bibentry}

\begin{document}

\maketitle

\begin{abstract}
Model inversion (MI) attacks have raised increasing concerns about privacy, which can reconstruct training data from public models. Indeed, MI attacks can be formalized as an optimization problem that seeks private data in a certain space. Recent MI attacks leverage a generative adversarial network (GAN) as an image prior to narrow the search space, and can successfully reconstruct even the high-dimensional data (e.g., face images). However, these generative MI attacks do not fully exploit the potential capabilities of the target model, still leading to a vague and coupled search space, i.e., different classes of images are coupled in the search space. Besides, the widely used cross-entropy loss in these attacks suffers from gradient vanishing. To address these problems, we propose Pseudo Label-Guided MI (PLG-MI) attack via conditional GAN (cGAN). At first, a top-\emph{n} selection strategy is proposed to provide pseudo-labels for public data, and use pseudo-labels to guide the training of the cGAN. In this way, the search space is decoupled for different classes of images. Then a max-margin loss is introduced to improve the search process on the subspace of a target class. Extensive experiments demonstrate that our PLG-MI attack significantly improves the attack success rate and visual quality for various datasets and models, notably, $2\sim3 \times$ better than state-of-the-art attacks under large distributional shifts.
Our code is available at: \textit{https://github.com/LetheSec/PLG-MI-Attack}.
\end{abstract}

\section{Introduction}\label{sec:intro}

Deep neural networks (DNNs) have revolutionized a wide variety of tasks, including computer vision, natural language processing, and healthcare. However, many practical applications of DNNs require training on private or sensitive datasets, such as facial recognition~\cite{taigman2014deepface} and medical diagnosis~\cite{rajpurkar2017chexnet}, which may pose some privacy threats. 
Indeed, the prior study of privacy attacks has demonstrated the possibility of exposing unauthorized information from access to a model~\cite{shokri2017membership, gopinath2019property, tramer2016stealing, fredrikson2015model}. 
In this paper, we mainly focus on model inversion (MI) attacks, a type of privacy attack that aims to recover the training data given a trained model. 
        
Typically, MI attacks can be formalized as an optimization problem with the goal of searching the input space for the sensitive feature value that achieves the highest likelihood under the target model. However, when attacking DNNs trained on more complex data (e.g., RGB images), directly solving the optimization problem via gradient descent tends to stuck in local minima, resulting in reconstructed images lacking clear semantic information. Recent work~\cite{zhang2020secret} proposed generative MI attacks (GMI), which used a generative adversarial network (GAN) to learn a generic prior of natural images, avoiding reconstructing private data directly from the unconstrained space. Generally, generative MI attacks can be summarized as the following two search stages:
\begin{itemize}
    \item\textbf{stage-1}: Generator Parameter Space Search. The adversary trains a generative model (i.e., searches for the optimal parameters) on a public dataset that only shares structural similarity with the private dataset.
    
    \item\textbf{stage-2}: Latent Vector Search. The adversary keeps searching the latent space of the generator trained in stage-1, until the output is close to the images in the private dataset.
\end{itemize} 
    
\begin{figure}
\centering

\includegraphics[width=1.0 \linewidth]{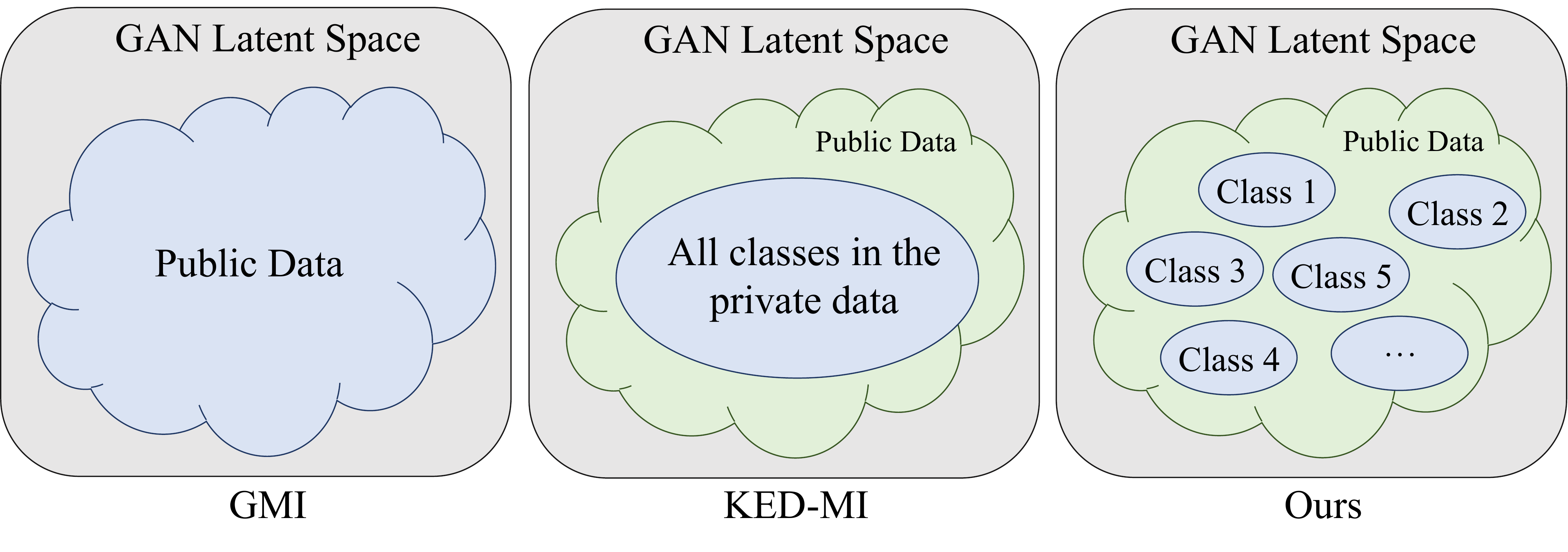}
\caption{Latent search space for different MI attacks. The blue area represents the latent space that needs to be searched to reconstruct a certain class of images. That is, the adversary should find an optimal latent point in the blue area, so that the generator outputs the private image of a specified class.}
\label{fig:search_space}
\end{figure}

Notably, GMI totally ignored the potential capability of the target model for the training process. Inspired by semi-supervised GAN~\cite{salimans2016improved}, KED-MI~\cite{chen2021knowledge} adopted a classifier as the GAN discriminator and utilized the target model to provide soft labels for public data during the training process in stage-1, which achieved the state-of-the-art MI attack performance. Even so, the attack performance is still underwhelming, especially when public and private data have a large distributional shift. We infer possible limitations of existing work as follows:
\begin{itemize}
    \item [1)] \textbf{Class-Coupled Latent Space.}
    The generator obtained by existing work in stage-1 is class-coupled. When the adversary reconstructs a specified class of target in stage-2, it needs to search in the latent space of all classes, which easily causes confusion of feature information between different classes (see~\Fref{fig:search_space}).
    \item [2)] \textbf{Indirectly Constrains on the Generator.}
    The KED-MI attack adopts the semi-supervised GAN framework, which indirectly constrains the generator with class information through the discriminator. However, this implicit constraint relies too much on the discriminator and lacks task specificity for MI.
    \item [3)]\textbf{Gradient Vanishing Problem.}
    Previous MI attacks have commonly adopted cross-entropy (CE) loss as the optimization goal in stage-2. However, the cross-entropy loss will cause the gradient to decrease and tend to vanish as the number of iterations increases, resulting in the search process to slow down or even stop early.
\end{itemize}

To address the above limitations, we propose a novel pseudo label-guided MI (PLG-MI) attack. Specifically, we first propose a simple but effective top-\emph{n} selection strategy, which can provide pseudo-labels for public data. Then we introduce the conditional GAN (cGAN) to MI attacks and use the pseudo-labels to guide the training process, enabling it to learn more specific and independent image distributions for each class. In addition, we also impose a task-specific explicit constraint directly on the generator so that class information can be embedded into the latent space. This constraint can force images to be generated towards specific classes in the private data. As shown in~\Fref{fig:search_space}, it can be considered as an approximate division of the GAN latent space into separate class subspaces. When reconstructing the private data of an arbitrary class in stage-2, only the corresponding subspace needs to be searched, which avoids confusion between different classes.

Our contributions can be summarized as follows:
\begin{itemize}
    \item We propose Pseudo Label-Guided MI (PLG-MI) attack, which can make full use of the target model and leverage pseudo-labels to guide the output of the generator during the training process.
        
    \item We propose a simple but effective strategy to provide public data with pseudo-labels, which can provide corresponding features according to specific classes in the private dataset.
     
    \item We demonstrate the gradient vanishing problem of cross-entropy loss commonly adopted in previous MI attacks and use max-margin loss to mitigate it.
        
    \item Extensive experiments demonstrate that the PLG-MI attack greatly boosts the MI attack and achieves state-of-the-art attack performance, especially in the presence of a large distributional shift between public and private data.
\end{itemize}
    
\begin{figure}
\centering
\includegraphics[width=1.0 \linewidth]{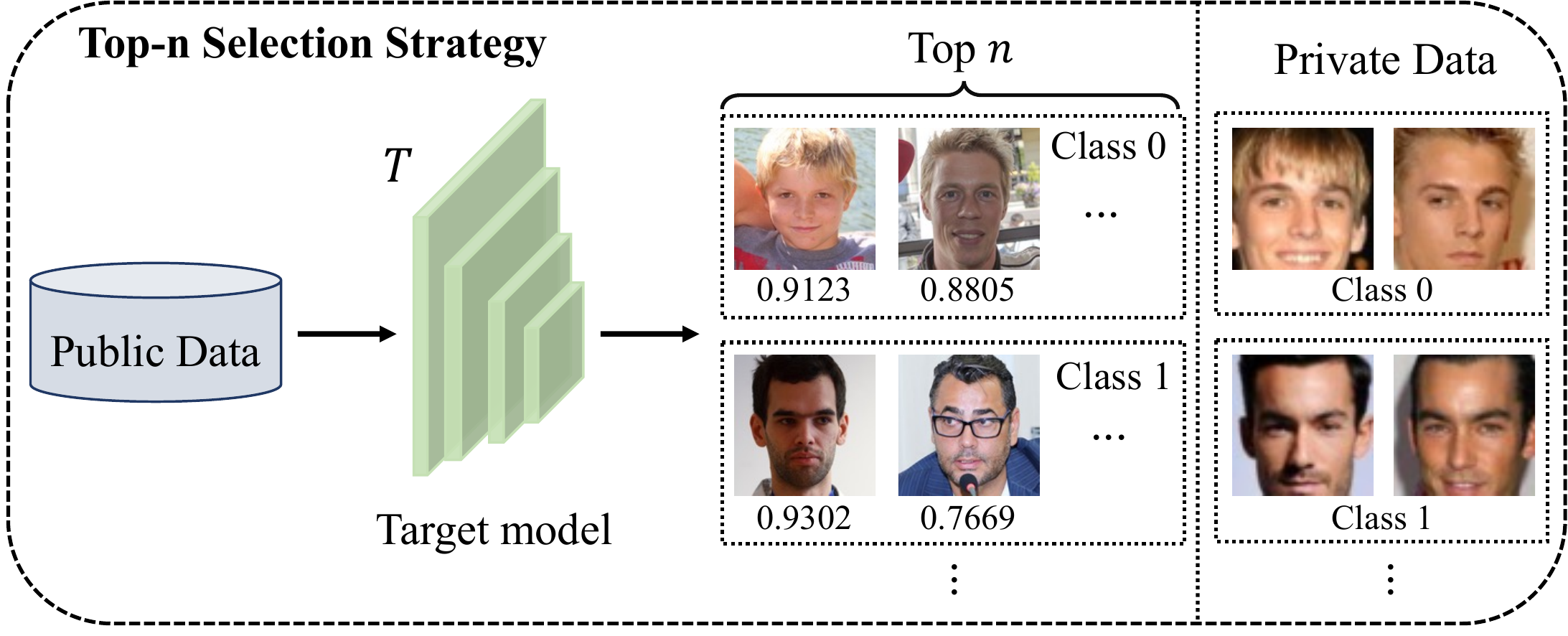}
\caption{Top-\emph{n} selection strategy. Input the images of public data into the target model and select the top $n$ images with the highest confidence for each class. The right half shows images of the corresponding class in the private dataset.}
\label{fig:top_n_selection}
\end{figure}
    
\section{Related Work}\label{sec:related_work}
	
\paragraph{Basic Model Inversion Attacks.} 
Fredrikson \etal~\cite{fredrikson2014privacy} first studied MI attacks in the context of genomic privacy and demonstrated that access to linear regression models for personalized medicine can be abused to recover private genomic properties of individuals in the training dataset. Fredrikson \etal~\cite{fredrikson2015model} later proposed an optimization algorithm based on gradient descent for MI attacks, which can recover grayscale face images from shallow networks. However, these basic MI attacks that reconstruct private data directly from the pixel space failed when the target models are DNNs.
	
\paragraph{Generative Model Inversion Attacks.} 
To make it possible to launch MI attacks against DNNs, Zhang \etal~\cite{zhang2020secret} proposed generative model inversion (GMI) attacks, which trained a GAN on public data as an image prior and then restricted the optimization problem in the latent space of the generator. Wang \etal~\cite{wang2021variational} proposed viewing MI attacks as a variational inference (VI) problem and provided a framework using deep normalizing flows in the extended latent space of a StyleGAN~\cite{karras2020analyzing}. Chen \etal~\cite{chen2021knowledge} adopted the semi-supervised GAN framework to improve the training process by including soft labels produced by the target model. Kahla \etal~\cite{kahla2022label} extended generative MI attacks to black-box scenarios where only hard labels are available.

\begin{figure*}
\centering
\includegraphics[width=1.0 \linewidth]{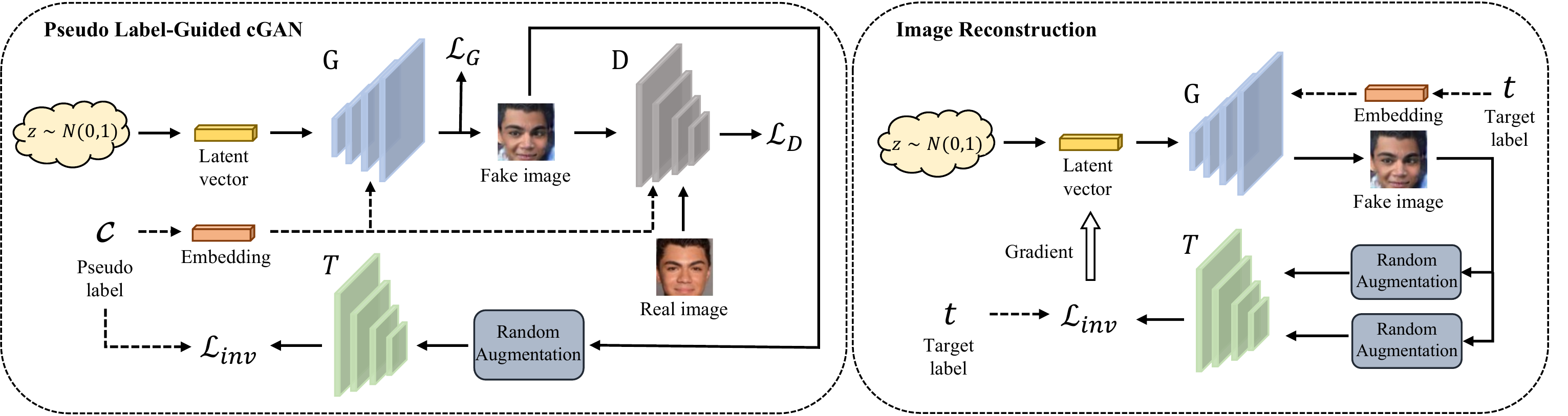}
\caption{The overall pipeline of the proposed two-stage model inversion attack algorithm. Stage 1: Train a conditional GAN on the public data with guidance provided by pseudo-labels and knowledge of the target model. Stage 2: Leverage the trained generator to reconstruct the specific class of private images using $\mathcal{L}_{inv}$.}
\label{fig:PLG_MI_framework}
\end{figure*}

\section{Method}\label{sec:method}
	
In this section, we will first discuss the threat model and then present our attack method in detail. 

\subsection{Threat Model}

\paragraph{Adversary’s Goal.}
Given a target model $T: [0,1]^{d} \rightarrow \mathbb{R}^{|C|}$ and an arbitrary class $c^{*} \in C$, the adversary aims to reconstruct a representative sample $x^{*}$ of the training data of the class $c^{*}$; $d$ represents the dimension of the model input; $C$ denotes the set of all class labels of the training data and $|C|$ is the size of the label set. It should be emphasized that the reconstructed data need to have good semantic information for human recognition. In this paper, we focus on the attack against face recognition models, that is, the adversary's goal is to reconstruct its corresponding face image from the target model according to the specified identity.
	
\paragraph{Adversary’s Knowledge.}
In this paper, we focus on white-box MI attacks, which means that the adversary can have access to all parameters of the target model. In addition, following the settings in previous work~\cite{zhang2020secret, chen2021knowledge, kahla2022label}, the adversary can gain a public dataset of the target task that only shares structural similarity with the private dataset without any intersecting classes. For example, the adversary knows the target model is for face recognition, he can easily leverage an existing open-sourced face dataset or crawl face images from the Internet as the public data.
	
\subsection{Pseudo-Labels Generation} 
	
\paragraph{Top-\emph{n} Selection Strategy.}
In order to make the public data have pseudo-labels to guide the training of the generator, we propose a top-\emph{n} selection strategy, as shown in \Fref{fig:top_n_selection}. This strategy aims to select the best matching $n$ images for each pseudo-label from public data. These pseudo-labels correspond to classes in the private dataset. Specifically, we feed all images in public data into the target model and get the corresponding prediction vectors. Then for a certain class $k$, we sort all images in descending order of $k_{th}$ value in their prediction vectors and select the top $n$ images to assign the pseudo-label $k$.

	
Generally, if the target model has high confidence in the $k_{th}$ class for a certain image, this image can be considered to contain discriminative features of the $k_{th}$ class. We define $F^{k}_{pri}$ and $F^{k}_{pub}$ as the distributions of discriminative features contained in the $k_{th}$ class of private data and pseudo-labeled public data. It can be inferred that $F^{k}_{pub}$ and $F^{k}_{pri}$ have intersection, which means that the adversary can obtain the required information by sufficiently searching $F^{k}_{pub}$, making the private information in $F^{k}_{pri}$ leaked more easily and accurately.




\paragraph{Narrow the Search Space via Pseudo-Labels.}
This strategy can narrow the search space of latent vectors in stage-2. Specifically, after reclassifying the public data, we can directly learn the feature distribution of images for each class. When reconstructing images of the $k_{th}$ class in the private dataset, it is only necessary to search for required features in $F^{k}_{pub}$, while reducing the interference of irrelevant features from $F^{i \neq k}_{pub}$. Taking face recognition as an example, suppose that the $k_{th}$ class of the private dataset is a young white man with blond hair, then the $k_{th}$ class of the pseudo-labeled public data is also mostly white people with blond hair, as shown in \Fref{fig:top_n_selection}. Consequently, the key features can be preserved and the useless features (e.g., other skin tones or hair colors, etc.) are eliminated, thereby narrowing the search space.
	
\begin{figure*}
\centering
\includegraphics[width=0.98 \linewidth]{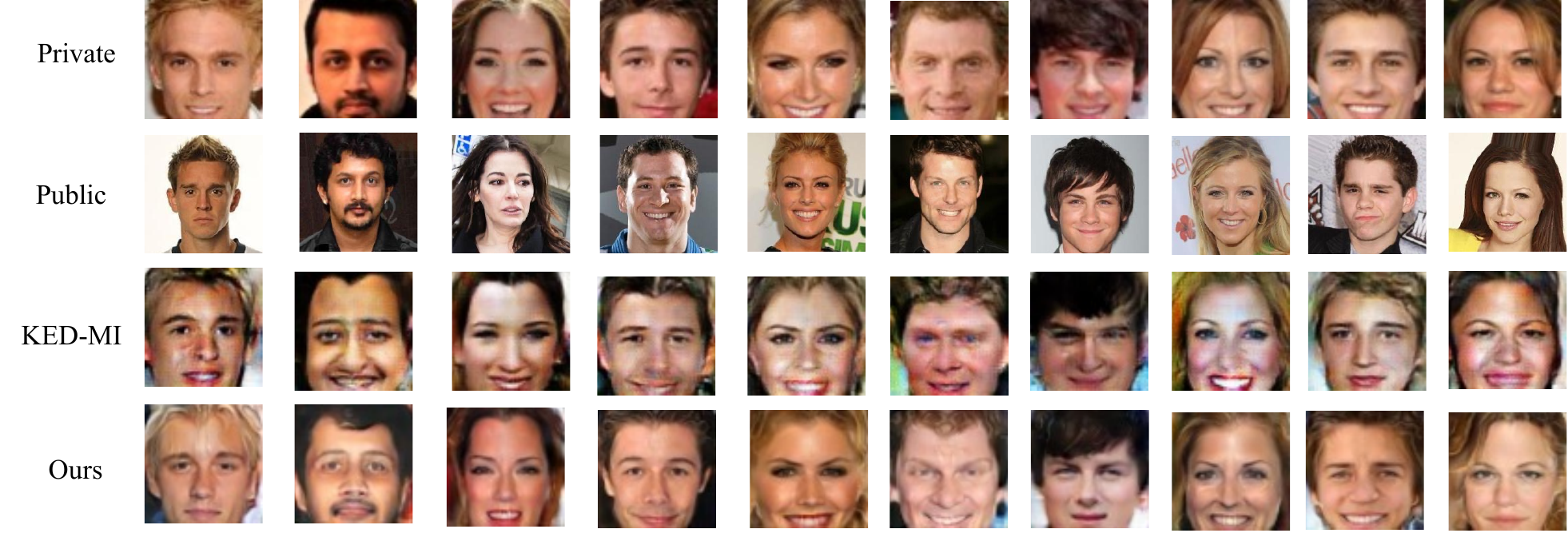}
\caption{Visual comparison for attacking VGG16 trained on CelebA. The first row shows ground truth images of target identity in the private data. The second row shows the images from the public data with the highest confidence in the target identity. The third and last rows demonstrate the reconstructed images of the target identity using KED-MI and our attack, respectively.}
\label{fig:celeba_images}
\end{figure*}

\subsection{Pseudo Label-Guided MI Attack}
    
An overview of our attack is illustrated in~\Fref{fig:PLG_MI_framework}, which consists of two stages. In stage-1, we train a conditional GAN on public data under the guidance of pseudo-labels. In stage-2, we use the trained generator to reconstruct private images of specified classes. 
	
\paragraph{Problem Formulation.} 
At first, we formulate the MI problem in the context of image classification (i.e., face recognition) with DNNs. We use $\mathcal{D}_{s}$ to denote the private dataset with sensitive information and $\mathcal{D}_{p}$ to denote the public dataset available to the adversary. Then using the top-\emph{n} selection strategy to obtain a pseudo-labeled public dataset $\mathcal{D}_{r}$. We denote a sample image as $x \in \mathcal{D}_s$, and its corresponding label as $y \in \{1,\dots,K\}$, where $K$ denotes the number of classes. Note that the original $\mathcal{D}_{p}$ does not have any class intersection with $\mathcal{D}_{s}$, while the $\mathcal{D}_{r}$ has the pseudo-labels $\tilde{y} \in \{1,\dots,K\}$. In the typical case, the target model $T$ will be trained on $\mathcal{D}_{s}$ to learn the mapping from the input space to the probability vectors. 

In generative MI attacks, the adversary uses $\mathcal{D}_p$ to train a GAN and then optimizes the input to the generator, instead of directly optimizing from the pixel space. Denote the trained generator by $G(z)$, where $z \sim \mathcal{N}(0, 1)$ is the latent vector. The optimization problem can be formulated as follows:
\begin{linenomath*}
\begin{equation}\label{eq:GMI}
z^{*} = \underset{\hat{z}}{\arg \min} \ \mathcal{L}_{inv}(T(G(\hat{z})), c),
\end{equation}
\end{linenomath*}
where $c$ is the target class in $\mathcal{D}_{s}$, and $\mathcal{L}_{inv}$ is a classification loss (e.g., cross-entropy). Then the reconstructed images can be obtained by $x^{*} = G(z^{*})$.

\paragraph{Pseudo Label-Guided cGAN.}
Although existing generative MI attacks~\cite{zhang2020secret, chen2021knowledge} can learn a prior of natural images, they do not take into account the possible effects of class labels, thus causing all classes to be coupled together in the latent space. This makes it difficult to directly search for private images of the specified class. As mentioned before, in order to narrow the search space and conduct a more independent latent search process. We propose to train a conditional GAN~\cite{miyato2018cgans} to model the feature distribution of each class and use pseudo-labels to guide the direction of the generated images.

Formally, for training the discriminator in the cGAN, we use a hinge version of the standard adversarial loss:
\begin{linenomath*}
\begin{equation}\label{eq:D_loss}
\begin{aligned}
\mathcal{L}_{D}=&E_{q(\tilde{y})}\left[E_{q(x \mid \tilde{y})}[\max (0,1-D(x, \tilde{y})]]+\right. \\
&E_{q(\tilde{y})}\left[E_{p(z)}[\max (0,1+D(G(z, \tilde{y}), \tilde{y}))],\right.
\end{aligned}
\end{equation}
\end{linenomath*}
where $q(\tilde{y})$ and $q(x|\tilde{y})$ are the pseudo-label distribution of \textbf{$\mathcal{D}_{r}$} and the image distribution in the corresponding class, respectively. $p(z)$ is standard Gaussian distribution and $G(z, \tilde{y})$ is the conditional generator.
	
To make the generated image more accurate, we use the pseudo-labels $\tilde{y}$ of $\mathcal{D}_r$ to impose an explicit constraint on the generator. The constraint, in principle, guides the generated images to belong to a certain class in the private dataset. Besides, we add a stochastic data augmentation module that performs random transformations on the generated images, including resizing, cropping, horizontal flipping, rotation, and color jittering. This module provides more stable convergence to realistic images while constraining. Then the loss function for the generator can be defined as:
\begin{linenomath*}
\begin{equation}\label{eq:G_loss}
\begin{aligned}
\mathcal{L}_{G} =& -E_{q(\tilde{y})}\left[E_{p(z)}\left[D(G(z, \tilde{y}), \tilde{y}))\right]\right] + \\
&\alpha \mathcal{L}_{inv}(T(\mathcal{A}(G(z, \tilde{y}))), \tilde{y}),
\end{aligned}
\end{equation}
\end{linenomath*}
where $T$ is the target model being attacked, $\mathcal{A}$ is a set of random augmentations, $\mathcal{L}_{inv}$ is the max-margin loss which we will introduce later, and $\alpha$ is a regularization coefficient.

\paragraph{Image Reconstruction.}
    
After getting the GAN trained on the public data, we can use it to reconstruct images of a specified class in the private dataset, as shown in the right half of~\Fref{fig:PLG_MI_framework}. Specifically, given a target class $c$, we aim to search for appropriate latent vectors, so that the generated images constantly approach the images in $c$. Since we use a conditional generator, only the subspace of the specified class needs to be searched. In order to ensure that reconstructed images are not deceptive (e.g., adversarial example) or just stuck in a local minimum, we transform generated images randomly resulting in multiple correlated views. Intuitively, if the reconstructed image truly reveals key discriminative features of the target class, its class should remain consistent across these views. The objective can be defined as follows:
\begin{linenomath*}
\begin{equation}\label{eq:Rec_Loss}
z^{*} = \underset{\hat{z}}{\arg \min}\  \sum_{i=1}^{m} \mathcal{L}_{inv}(T( \mathcal{A}_{i}(G(\hat{z}, c))),c),
\end{equation}
\end{linenomath*}
where $\hat{z}$ is the latent vector to be optimized, $\mathcal{L}_{inv}$ is the max-margin loss, $\mathcal{A}_{i}$ is a set of random data augmentations, and $m$ is the number of augmented views. Then we can obtain the reconstructed images by $x^{*} = G(z^{*},c)$.

\begin{table*}[ht!]
\centering
\resizebox{1.0\textwidth}{!}{
    \begin{tabular}{@{}cccccccccc@{}}
    \toprule
                                & \multicolumn{3}{c}{\textbf{VGG16}}                  & \multicolumn{3}{c}{\textbf{ResNet-152}}            & \multicolumn{3}{c}{\textbf{Face.evoLVe}}             \\
                                & \textbf{GMI} & \textbf{KED-MI} & \textbf{Ours}      & \textbf{GMI} & \textbf{KED-MI} & \textbf{Ours}     & \textbf{GMI} & \textbf{KED-MI} & \textbf{Ours}       \\ \midrule
    \textbf{Attack Acc $\uparrow$}       & .21$\pm$.0028    & .63$\pm$.0018       & \textbf{.97$\pm$.0001} & .31$\pm$.0035    & .74$\pm$.0028       & \textbf{1.$\pm$.0000} & .29$\pm$.0030    & .74$\pm$.0013       & \textbf{.99$\pm$.0001} \\
    \textbf{Top-5 Attack Acc $\uparrow$} & .42$\pm$.0021    & .87$\pm$.0015       & \textbf{1.$\pm$.0000}  & .55$\pm$.0045    & .93$\pm$.0006       & \textbf{1.$\pm$.0000} & .54$\pm$.0040    & .94$\pm$.0009       & \textbf{1.$\pm$.0000}   \\
    \textbf{KNN Dist $\downarrow$}         & 1712.57      & 1391.52         & \textbf{1120.61}   & 1630.25      & 1323.16         & \textbf{1026.71}  & 1638.94      & 1310.15         & \textbf{1103.03}    \\
    \textbf{FID $\downarrow$}               & 42.86        & 30.92           & \textbf{18.63}     & 42.50        & 26.23           & \textbf{23.22}    & 41.53        & 27.92           & \textbf{26.75}      \\ \bottomrule
    \end{tabular}
    }
\caption{Attack performance comparison on various models trained on CelebA. $\uparrow$ and $\downarrow$ respectively symbolize that higher and lower scores give better attack performance.}
\label{tab: celeba_results}
\end{table*}
    
    
\begin{table*}[ht!]
\centering
\resizebox{1.0\textwidth}{!}{
    \begin{tabular}{@{}cccccccccc@{}}
    \toprule
    \multicolumn{2}{c}{\multirow{2}{*}{\textbf{}}}          & \multicolumn{4}{c}{\textbf{FFHQ $\rightarrow$ CelebA}}                                                                          & \multicolumn{4}{c}{\textbf{FaceScrub $\rightarrow$ CelebA}}                                                                   \\
    \multicolumn{2}{c}{}                                    & \textbf{Attack Acc   $\uparrow$} & \textbf{Attack Acc 5 $\uparrow$} & \textbf{KNN Dist $\downarrow$} & \textbf{FID$\downarrow$} & \textbf{Attack Acc $\uparrow$} & \textbf{Attack Acc 5 $\uparrow$} & \textbf{KNN Dist$\downarrow$} & \textbf{FID$\downarrow$} \\ \midrule
    \multirow{3}{*}{\textbf{VGG16}}       & \textbf{GMI}    & .11$\pm$.0009                        & .27$\pm$.0048                        & 1771.34                        & 57.05                    & .02$\pm$.0004                      & .07$\pm$.0008                         & 1997.16                       & 150.19                   \\
                                          & \textbf{KED-MI} & .34$\pm$.0026                        & .62$\pm$.0015                        & 1555.57                        & 49.51                    & .05$\pm$.0008                      & .14$\pm$.0006                         & 1772.85                       & 97.56                    \\
                                          & \textbf{Ours}   & \textbf{.89$\pm$.0006}               & \textbf{.97$\pm$.0002}               & \textbf{1284.16}               & \textbf{27.32}           & \textbf{.55$\pm$.0020}             & \textbf{.77$\pm$.0012}                & \textbf{1474.22}              & \textbf{27.99}           \\ \midrule
    \multirow{3}{*}{\textbf{Face.evoLVe}} & \textbf{GMI}    & .13$\pm$.0009                        & .31$\pm$.0028                        & 1739.88                        & 56.66                    & .03$\pm$.0004                      & .10$\pm$.0012                         & 1918.40                        & 112.96                   \\
                                          & \textbf{KED-MI} & .47$\pm$.0021                        & .74$\pm$.0013                        & 1489.67                        & 44.48                    & .09$\pm$.0006                      & .24$\pm$.0019                         & 1712.31                       & 99.78                    \\
                                          & \textbf{Ours}   & \textbf{.95$\pm$.0004}               & \textbf{.99$\pm$.0001}               & \textbf{1241.41}               & \textbf{25.57}           & \textbf{.57$\pm$.0013}             & \textbf{.78$\pm$.0012}                & \textbf{1502.82}              & \textbf{34.10}           \\ \midrule
    \multirow{3}{*}{\textbf{ResNet-152}}  & \textbf{GMI}    & .17$\pm$.0026                        & .37$\pm$.0030                        & 1687.82                        & 47.11                    & .04$\pm$.0011                      & .14$\pm$.0020                         & 1865.44                       & 109.16                   \\
                                          & \textbf{KED-MI} & .74$\pm$.0028                        & .93$\pm$.0006                        & 1323.16                        & 26.23                    & .15$\pm$.0011                      & .36$\pm$.0020                         & 1636.81                       & 72.72                    \\
                                          & \textbf{Ours}   & \textbf{1.$\pm$.0000}                & \textbf{1.$\pm$.0000}                & \textbf{1026.71}               & \textbf{23.22}           & \textbf{.68$\pm$.0020}             & \textbf{.87$\pm$.0011}                & \textbf{1360.67}              & \textbf{27.49}           \\ \bottomrule
\end{tabular}
}
\caption{Attack performance comparison in the presence of a large distributional shift between public and private data. $A \rightarrow B$ represents the GAN and target model trained on datasets A and B, respectively.}
\label{tab: cross_dataset_results}
\end{table*}

\subsection{A Better Loss for MI Attacks}
	
\paragraph{Gradient Vanishing Problem.}
Existing MI attacks have commonly adopted the cross-entropy (CE) loss as $\mathcal{L}_{inv}$. During attack optimization, the CE loss will cause the gradient to decrease and tend to vanish as the number of iterations is increased. 
For the target class $c$, the derivative of cross-entropy loss $\mathcal{L}_{\text{CE}}$ with respect to the output logits $\mathbf{o}$ can be derived as (see Appendix for derivation):
\begin{linenomath*}
\begin{equation}\label{eq:CE_loss_derivative}
\frac{\partial  \mathcal{L}_{\text{CE}}}{\partial \mathbf{o}}=\mathbf{p}-\mathbf{y}_{c}. \\
\end{equation}
\end{linenomath*}
Here, $\mathbf{p}$ is the probability vector of the softmax output, that is $\mathbf{p}=\left[p_{1}, p_{2}, \dots ,p_{K}\right]$, $p_c \in [0,1]$ denotes the predicted probability of class $c$. $\mathbf{y}_{c}$ is the one-hot encoded vector of class $c$, that is, $\mathbf{y}_{c}=\left[0_{1}, \dots, 1_{c}, \dots ,0_{K}\right]$. Then,  \Eref{eq:CE_loss_derivative} can be rewritten as:
\begin{linenomath*}
\begin{equation}\label{eq:CE_loss_derivative_2}
\frac{\partial  \mathcal{L}_{\text{CE}}}{\partial \mathbf{o}} =\left[p_{1}, \dots, p_{c}-1, \dots p_{K}\right].
\end{equation}
\end{linenomath*}
According to \Eref{eq:CE_loss_derivative_2}, as the generated image gradually approaches the target class during optimization, $p_c$ will quickly reach 1 while $p_{i \neq c}$ will continue to decrease to 0. Eventually, this changing trend will cause the gradient of $\mathcal{L}_\text{CE}$ to vanish, making it difficult to search the latent vector of the generator.

\paragraph{Max-Margin Loss.}
To address this problem, we propose to replace the CE loss with the max-margin loss, which has been used in adversarial attacks to produce stronger attacks~\cite{carlini2017towards, sriramanan2020guided}. In addition, we eliminate the softmax function and optimize the loss directly on the logits. The max-margin loss $\mathcal{L}_\text{MM}$ we use as $\mathcal{L}_{inv}$ is as follows:
\begin{linenomath*}
\begin{equation}\label{eq:maximum_margin_loss}
\mathcal{L}_\text{MM} = -l_{c}(x) + \max_{j \neq c}l_{j}(x),
\end{equation}
\end{linenomath*}
where $l_{c}$ denotes the logit with respect to the target class $c$. For the target class $c$, the derivative of $\mathcal{L}_\text{MM}$ with respect to the logits can be derived as (see Appendix for derivation):
\begin{linenomath*}
\begin{equation}
\frac{\partial \mathcal{L}_\text{MM}}{\partial \mathbf{o}}=\mathbf{y}_{j}-\mathbf{y}_{t},
\end{equation}
\end{linenomath*}
where $\mathbf{y}_{j}$ and $\mathbf{y}_{t}$ represent one-hot encoded vectors, so the elements of the gradient consist of constants, thus avoiding gradient vanishing problem. Moreover, max-margin loss encourages the algorithm to find the most representative sample in the target class while also distinguishing it from other classes. Compared with the cross-entropy loss, max-margin loss is more in line with the goal of MI attacks (further comparisons are given in the Appendix).

\section{Experiments}\label{sec:experiment}
In this section, we first provide a detailed introduction of the experimental settings. To demonstrate the effectiveness of our methods, we evaluate the proposed PLG-MI attack from several perspectives. The baselines that we will compare against are GMI proposed in~\cite{zhang2020secret} and KED-MI proposed in~\cite{chen2021knowledge}, the latter achieved the state-of-the-art result for attacking DNNs.

\begin{table*}[ht!]
\centering
\resizebox{1.0\textwidth}{!}{
\begin{tabular}{@{}ccccccccccc@{}}
\toprule
\multicolumn{2}{c}{\multirow{2}{*}{\textbf{FaceScrub $\rightarrow$ CelebA}}} & \multicolumn{3}{c}{\textbf{VGG16}}                                                            & \multicolumn{3}{c}{\textbf{Face.evoLVe}}                                                    & \multicolumn{3}{c}{\textbf{ResNet-152}}                                                     \\
    \multicolumn{2}{c}{}                                                         & \textbf{Attack Acc   $\uparrow$} & \textbf{KNN Dist $\downarrow$} & \textbf{FID $\downarrow$} & \textbf{Attack Acc $\uparrow$} & \textbf{KNN Dist $\downarrow$} & \textbf{FID $\downarrow$} & \textbf{Attack Acc $\uparrow$} & \textbf{KNN Dist $\downarrow$} & \textbf{FID $\downarrow$} \\ \midrule
    \multirow{2}{*}{\textbf{VGG16}}                  & \textbf{KED-MI}           & .05$\pm$.0008                        & 1772.85                        & 97.56                     & .10$\pm$.0006                      & 1694.13                        & 87.79                     & .12$\pm$.0009                      & 1638.34                        & 87.24                     \\
                                                        & \textbf{Ours}             & \textbf{.55$\pm$.0020}               & \textbf{1474.22}               & \textbf{27.99}            & \textbf{.76$\pm$.0017}             & \textbf{1356.23}               & \textbf{25.57}            & \textbf{.81$\pm$.0016}             & \textbf{1282.36}               & \textbf{23.74}            \\ \midrule
    \multirow{2}{*}{\textbf{Face.evoLVe}}            & \textbf{KED-MI}           & .05$\pm$.0006                        & 1773.14                        & 103.00                    & .09$\pm$.0006                      & 1712.31                        & 99.78                     & .13$\pm$.0018                      & 1646.85                        & 96.04                     \\
                                                        & \textbf{Ours}             & \textbf{.54$\pm$.0019}               & \textbf{1472.86}               & \textbf{31.16}            & \textbf{.57$\pm$.0013}             & \textbf{1502.82}               & \textbf{34.10}            & \textbf{.71$\pm$.0015}             & \textbf{1390.84}               & \textbf{27.84}            \\ \midrule
    \multirow{2}{*}{\textbf{ResNet-152}}             & \textbf{KED-MI}           & .06$\pm$.0004                        & 1776.78                        & 107.75                    & .12$\pm$.0012                      & 1697.37                        & 88.28                     & .15$\pm$.0011                      & 1636.81                        & 72.72                     \\
                                                        & \textbf{Ours}             & \textbf{.57$\pm$.0019}               & \textbf{1427.91}               & \textbf{28.35}            & \textbf{.68$\pm$.0026}             & \textbf{1385.99}               & \textbf{27.30}            & \textbf{.68$\pm$.0020}             & \textbf{1360.67}               & \textbf{27.49}            \\ \bottomrule
\end{tabular}
}
\caption{Attack performance comparison when using models with different architectures in the GAN training stage and the image reconstruction stage. The public dataset is FaceScrub which has a larger distributional shift with CelebA.}
\label{tab: cross_model_dataset_results2}
\end{table*}

\subsection{Experimental Setting}
	
\paragraph{Datasets.}
For face recognition, we select three widely used datasets for experiments: CelebA~\cite{liu2015deep}, FFHQ~\cite{karras2019style} and FaceScrub~\cite{ng2014data}. CelebA contains 202,599 face images of 10,177 identities with coarse alignment. FFHQ consists of 70,000 high-quality PNG images and contains considerable variation in terms of age, ethnicity and image background. FaceScrub is a dataset of URLs for 100,000 images of 530 individuals. Similar to previous work~\cite{zhang2020secret, chen2021knowledge, kahla2022label}, we crop the images of all datasets at the center and resize them to $64 \times 64$. More experiments on MNIST~\cite{lecun1998gradient}, CIFAR-10~\cite{krizhevsky2009learning} and ChestX-Ray~\cite{wang2017chestx} can be found in the Appendix.

\paragraph{Models.}
Following the setting of the state-of-the-art MI attack~\cite{chen2021knowledge}, we evaluate our attack on three deep models with various architectures: (1) VGG16~\cite{simonyan2014very}; (2) Face.evoLVe~\cite{cheng2017know}; and (3) ResNet-152~\cite{he2016deep}.

\paragraph{Implementation Details.}
In the standard setting, previous MI attacks usually split the dataset into two disjoint parts: one part used as the private dataset to train the target model and the other as the public dataset. For training the target model, we use 30,027 images of 1,000 identities from CelebA as the private dataset. The disjoint part of CelebA is used to train the generator. However, this setting is too easy under our stronger PLG-MI attack. Therefore we focus on the scenario where the public dataset has a larger distributional shift with the private dataset, i.e., using two completely different datasets. Specifically, we use FFHQ and FaceScrub as public datasets respectively to train the generator. We set $n=30$ for the top-\emph{n} selection strategy, that is, each public dataset consists of 30,000 selected images that are reclassified into 1,000 classes by pseudo-labels. In stage-1, the GAN architecture we use is based on~\cite{miyato2018cgans}. We apply spectral normalization~\cite{miyato2018spectral} to the all of the weights of the discriminator to regularize the Lipschitz constant. To train the GAN, we used Adam optimizer with a learning rate of $0.0002$, a batch size of $64$ and $\beta=(0, 0.9)$. The hyperparameter $\alpha$ in~\Eref{eq:G_loss} is set to $0.2$. In stage-2, we use the Adam optimizer with a learning rate of $0.1$ and $\beta = (0.9, 0.999)$. The input vector $z$ of the generator is drawn from a zero-mean unit-variance Gaussian distribution. We randomly initialize $z$ for $5$ times and optimize each round for $600$ iterations.

\subsection{Evaluation Metrics}
The evaluation of the MI attack is based on the similarity of the reconstructed image and the target class image in the human-recognizable features. In line with previous work, we conducted both qualitative evaluation through visual inspection as well as quantitative evaluation. Specifically, the evaluation metrics we used are as follows.

\paragraph{Attack Accuracy (Attack Acc).} 
We first build an evaluation model, which has a different architecture from the target model. We then use the evaluation model to compute the top-1 and top-5 accuracy of the reconstructed image on the target class. Actually, the evaluation model can be viewed as a proxy for a human observer to judge whether a reconstruction captures sensitive information. We use the model in~\cite{cheng2017know} for evaluation, which is pretrained on MS-Celeb1M~\cite{guo2016ms} and then fine-tuned on training data of the target model.

\paragraph{K-Nearest Neighbor Distance (KNN Dist).}
Given a target class, we computed the shortest feature distance from a reconstructed image to the private images. The distance is measured by the $\ell_{2}$ distance between two images in the feature space, i.e., the output of the penultimate layer of the evaluation model. A lower value indicates that the reconstructed image is closer to the private data.
    
\begin{figure*}
\centering 
\includegraphics[width=1.0 \linewidth]{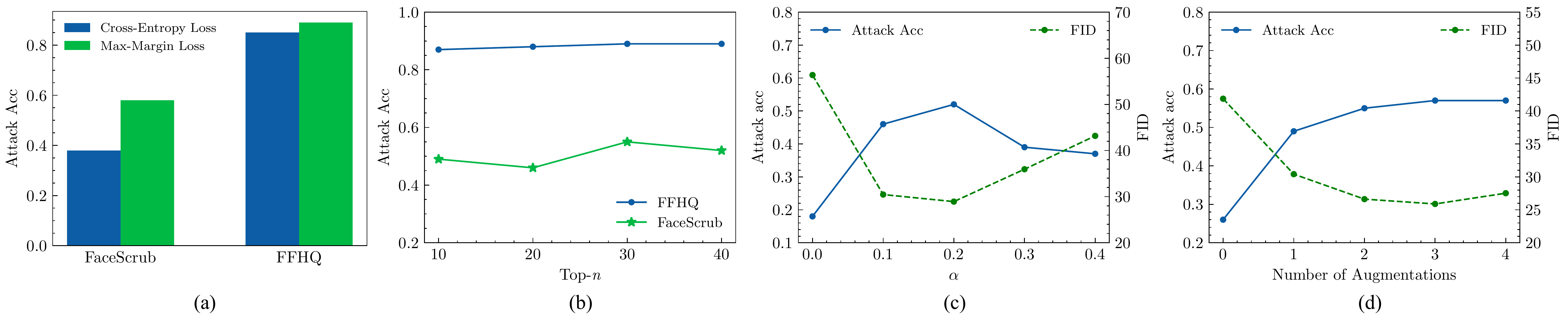}
\caption{(a)-(b) Attack performance with different $\mathcal{L}_{inv}$ and top-$n$ selection strategy, respectively. (c) Attack Acc and FID with constraints of various strengths in~\Eref{eq:G_loss}. (d) Attack Acc and FID with various numbers of augmentations in~\Eref{eq:Rec_Loss}. (c) and (d) use FaceScrub as the public dataset.}
\label{fig:ablation}
\end{figure*}
    
\paragraph{Fr\'echet Inception Distance (FID).}
FID~\cite{heusel2017gans} is commonly used in the work of GAN to evaluate the generated images. Lower FID values indicate that the reconstructed images have better quality and diversity, making it easier for humans to identify sensitive features in them. Following the baseline setting, we only compute FID values on images where the attack is successful.

\subsection{Experimental Results}
        	
\paragraph{Standard Setting.}
We first study the standard setting, i.e.,, dividing the CelebA dataset into a private dataset and a public dataset. As shown in~\Tref{tab: celeba_results}, our method outperforms the baselines on all three models. The reconstructed images produced by our PLG-MI attack can achieve almost 100\% accuracy on the evaluation model (i.e., Attack Acc), which outperforms the state-of-the-art method on average by approximately 30\%. Our method is also vastly superior in FID and KNN Dist, significantly improving the visual quality and the similarity of reconstructed images to private datasets.
	
\Fref{fig:celeba_images} shows the visual comparison of different methods. Compared with KED-MI, for different identities in the private dataset, our reconstructed images not only have more similar semantic features to the ground truth images, but are also more realistic. The second row shows the images of the public data with the highest confidence in the corresponding identity of the target model. We use the confidence as the basis for providing pseudo-labels in the top-\emph{n} selection strategy. In most cases, the images selected by the strategy can provide some of the characteristic features needed to reconstruct the identity, such as gender, hair color and skin tone. But in terms of details, it can still be easily distinguished that they are not the same identity. Therefore, the optimization process of stage-2 is also very critical, and these candidate features will be further filtered and reorganized to increasingly resemble the target identity.

\paragraph{Larger Distributional Shifts.}
To further explore scenarios where the baselines performed poorly, we conducted experiments in the setting with larger distributional shifts. As shown in ~\Tref{tab: cross_dataset_results}, GMI and KED-MI achieve the Attack Acc of 17\% and 74\% when exploiting FFHQ to attack ResNet-152, respectively. Since they do not fully explore the capability in the target model, the performance is poor in this more difficult setting. Compared with them, our PLG-MI attack achieves 89\% Attack Acc when attacking VGG16 using FFHQ as public data, which is about 2.5 times higher than KED-MI. And the Attack Acc reaches more than 95\% when attacking both Face.evoLVe and ResNet-152. In addition, there is a significant improvement in the FID for evaluating visual quality (e.g., a reduction from 49.51 to 27.32).
    
It should be emphasized that in the case of using FaceScrub as the public dataset, both GMI and KED-MI only obtain an almost failed attack performance. However, our method still maintains a high attack success rate, on average 50\% higher than KED-MI on Attack Acc. As for the FID, our method maintains a low value in all cases, with an average reduction of about 3 times relative to KED-MI. The lower KNN Dist also shows that our method can reconstruct more accurate private images. 
    
The results of FFHQ as public data are generally better than those of FaceScrub. The possible reason is that FaceScrub consists of face images crawled from the Internet, which is more different from the distribution of CelebA. Furthermore, we find that the degree of privacy leakage under MI attacks varies between model architectures. In our experiments, ResNet-152 is more likely to leak information in private datasets compared to VGG16 and Face.evoLVe. This phenomenon deserves further study in future work.
    
\paragraph{Generality of the GAN.}
The adversary may simultaneously perform MI attacks against multiple available target models to verify the correctness of the reconstructed image for the target identity. However, it will be very time-consuming to train a GAN separately for a specific target model each time. Therefore, we explore a new scenario where the architecture of the classifier used to help train the GAN is different from that of the target model.

\Tref{tab: cross_model_dataset_results2} compares the results of our method with those of the baselines. Our method yields clearly superior results for all three models under various evaluation metrics. When using VGG16 to provide prior information to train a generator in stage-1, and using it to attack ResNet-152 in stage-2, our method achieves 81\% Attack Acc, while KED-MI only achieves 12\%. In addition, the FID of KED-MI is approximately 3 times higher on average than ours. 
    
The experimental results of this scenario illustrate that regardless of the architecture of the models, our method is able to extract general knowledge to help recover sensitive information. Specifically, the generality of our cGAN can greatly improve the efficiency of attacks when there are multiple target models to be attacked. 


\paragraph{Ablation study.}
We further investigate the effects of the various components and hyperparameters in PLG-MI and conduct attacks against VGG16 trained on CelebA using FaceScrub or FFHQ as the public dataset. As shown in~\Fref{fig:ablation} (a), we compare the impact of different $\mathcal{L}_{inv}$ used in MI attacks. Using the max-margin loss brings a significant improvement on Attack Acc, especially when the distributional shift is larger (i.e., the public dataset is FaceScrub). \Fref{fig:ablation} (b) presents the results of using different $n$ in the top-\emph{n} selection strategy, its value has no significant impact on the attack performance, indicating that the strategy is not sensitive to $n$. \Fref{fig:ablation} (c) shows the attack performance and visual quality when imposing explicit constraints of various strengths on the generator and $\alpha=0.2$ is optimal among them. \Fref{fig:ablation} (d) shows that the data augmentation module in stage-2 has a great influence on improving the Attack Acc and the visual quality of reconstructed images. However, when the number of augmentations is greater than 2, the improvement in attack performance is small, but the attack efficiency will be reduced. Thus we finally force the reconstructed images to maintain identity consistency after 2 random augmentations.

\section{Conclusion} 
In this paper, we propose a novel MI attack method, namely PLG-MI attack. We introduce the conditional GAN and use pseudo-labels provided by the proposed top-\emph{n} selection strategy to guide the training process. In this way, the search space in the stage of image reconstruction can be limited to the subspace of the target class, avoiding the interference of other irrelevant features. Moreover, we propose to use max-margin loss to overcome the problem of gradient vanishing. Experiments show that our method achieves the state-of-the-art attack performance on different scenarios with various model architectures. For future work, the black-box MI attack is still in its infancy, and the idea of our method can be transferred to the black-box scenario to further improve its performance.

\section*{Acknowledgements}
This work was supported in part by the Natural Science Foundation of China
under Grant  U20B2047, 62102386, 62072421, 62002334 and 62121002.

\bibliography{main}

\clearpage 

\appendix

\section{Appendix}
\subsection{The Derivatives of Cross-Entropy Loss and Max-Margin Loss }
\paragraph{Cross-Entropy Loss}\label{appendix:CE_loss}

We define the CE loss for the \emph{K}-
classification task as follows:
\begin{linenomath*}
\begin{equation}\label{appendix_eq:ce_loss}
    \mathcal{L}=-\sum_{i=1}^{K} y_i \log \left(p_{i}\right),
\end{equation}
\end{linenomath*}
where $\mathbf{y}_i$ denotes the one-hot encoded vector of class $i$, in which only the $i_{th}$ element $y_i$ is 1, and the rest are 0. (i.e., $\mathbf{y}_i=[0_1,\dots,1_i,\dots,0_K]$). Thus, for the specified target class $t$, \Eref{appendix_eq:ce_loss} can be rewritten as:
\begin{linenomath*}
\begin{equation}\label{appendix_eq:ce_loss2}
    \mathcal{L}= -\log(p_t).
\end{equation}
\end{linenomath*}
Here, we denote the logits of the model output as  $\mathbf{o}=[o_1,\dots,o_t,\dots,o_K]$, and denote the softmax output (i.e., probability vector) as $\mathbf{p}=[p_1,\dots,p_t,\dots,p_K]$. Then, the derivative of the CE Loss ($\mathcal{L}$) with respect to the logits ($\mathbf{o}$) is as follow:
\begin{linenomath*}
\begin{equation}\label{appendix_eq:ce_loss_derivatives}
    \frac{\partial \mathcal{L}}{\partial \mathbf{o}}=\frac{\partial \mathcal{L}}{\partial{\mathbf{p}}}\frac{\partial \mathbf{p}}{\partial{\mathbf{o}}}.
\end{equation}
\end{linenomath*}
It can be seen from \Eref{appendix_eq:ce_loss2} that $\mathcal{L}$ is only related to $p_t$, so we can get:
\begin{linenomath*}
\begin{equation}
    \frac{\partial \mathcal{L}}{\partial \mathbf{p}}=\left[0_{1},\dots,-\frac{1}{p_t},\dots,0_{K}\right].
\end{equation}
\end{linenomath*}
And $\frac{\partial \mathbf{p}}{\partial \mathbf{o}}$ of \Eref{appendix_eq:ce_loss_derivatives} is a Jacobian matrix, as follows:
\begin{linenomath*}
\begin{equation}\label{appendix_eq:Jacobian_matrix}
    \frac{\partial \boldsymbol{p}}{\partial \boldsymbol{o}}=\left(\begin{array}{ccc}
    \frac{\partial p_{1}}{\partial o_{1}} & \frac{\partial p_{1}}{\partial o_{2}} & \cdots \frac{\partial p_{1}}{\partial o_{K}} \\
    \vdots & \vdots & \ddots \\
    \frac{\partial p_{j}}{\partial o_{1}} & \frac{\partial p_{j}}{\partial o_{2}} & \cdots \frac{\partial p_{j}}{\partial o_{K}} \\
    \vdots & \vdots & \ddots \\
    \frac{\partial p_{K}}{\partial o_{1}} & \frac{\partial p_{K}}{\partial o_{2}} & \cdots \frac{\partial p_{K}}{\partial o_{K}}
    \end{array}\right).
\end{equation}
\end{linenomath*}
As only the element in row $t$ is not $0$ in \Eref{appendix_eq:Jacobian_matrix}, we arrive at:
\begin{linenomath*}
\begin{equation}\label{appendix_eq:ce_loss_derivatives_2}
\begin{aligned}
    &\frac{\partial \mathcal{L}_{C E}}{\partial \boldsymbol{o}}=\frac{\partial \mathcal{L}_{C E}}{\partial \boldsymbol{p}} \frac{\partial \boldsymbol{p}}{\partial \boldsymbol{o}}=-\frac{1}{p_{t}} \frac{\partial p_{t}}{\partial \boldsymbol{o}}\\
    &\text{where}~~p_{t}=\frac{e^{o_{j}}}{\sum_{i=1}^{K} e^{o_{i}}}.
\end{aligned}
\end{equation}
\end{linenomath*}
With \Eref{appendix_eq:ce_loss_derivatives_2}, for the case where $i \neq t$, the derivative of CE loss is as follows:
\begin{linenomath*}
\begin{equation}\label{appendix_eq:ce_loss_derivatives_3}
\begin{aligned}
    \frac{\partial L}{\partial o_{i}} 
    &=\frac{\partial L}{\partial p_{t}}\frac{\partial p_{t}}{\partial o_{i}} \\
    &=\frac{\partial L}{\partial p_{t}} \frac{0-e^{o_{t}} e^{o_{i}}}{\left(\sum_{i=1}^{K} e^{o_{i}}\right)^{2}} \\
    &=(-\frac{1}{p_{t}})(-p_{t} p_{i}) \\
    &=p_{i}.
\end{aligned}
\end{equation}
\end{linenomath*}
And for the case where $i = t$, the derivative of CE loss is as follows:
\begin{linenomath*} 
\begin{equation}\label{appendix_eq:ce_loss_derivatives_4}
\begin{aligned}
    \frac{\partial L}{\partial o_{i}} &=\frac{\partial L}{\partial p_{t}}\frac{\partial p_{t}}{\partial o_{i}} \\
    &=\frac{\partial L}{\partial p_{t}}\frac{e^{o_{t}} \sum_{i=1}^{K} e^{o_{i}}-e^{o_{t}} e^{o_{t}}}{\left(\sum_{i=1}^{K} e^{o_{i}}\right)^{2}} \\
    &=(-\frac{1}{p_{t}})\left(p_{t}-p_{t}^{2}\right) \\
    &=p_{t}-1.
\end{aligned}
\end{equation}
\end{linenomath*}
From \Eref{appendix_eq:ce_loss_derivatives_3} and \Eref{appendix_eq:ce_loss_derivatives_4}, we derive the final derivative of CE loss as:
\begin{linenomath*} 
\begin{equation}
\begin{aligned}
    \frac{\partial L}{\partial \mathbf{o}} &=\left[p_{1}, \dots, p_{t}-1, \dots p_{K}\right] \\
    &=\left[p_{1}, \dots ,p_{t}, \dots p_{K}\right]-\left[0_{1}, \dots ,1_{t}, \dots ,0_{K}\right] \\
    &=\mathbf{p}-\mathbf{y}_{t}.
\end{aligned}
\end{equation}
\end{linenomath*} 
\paragraph{Max-Margin Loss.}\label{appendix:MM_loss}
Without loss of generality, we can define the max-margin loss for a target class $t$ as follows:
\begin{linenomath*}
\begin{equation}
    \mathcal{L} = o_{j} - o_{t},
\end{equation}
\end{linenomath*}
where $j=\arg \max_{i \neq t}o_{i}$ indicates the highest class except for the target class $t$. We denote the derivative of $\mathcal{L}$ to the logits $\mathbf{o}$ as $\frac{\partial \mathcal{L}}{\partial \mathbf{o}}$. It can be deduced that $ \frac{\partial \mathcal{L}}{\partial o_{i}}=0$ when $i \neq j$ and $i \neq t$, and $\frac{\partial \mathcal{L}}{\partial o_{i}}$ is 1 and $-1$ when $i=j$ and $i=t$, respectively. Thus, we finally derive at:
\begin{linenomath*}
\begin{equation}
\frac{\partial L}{\partial \mathbf{o}}=\mathbf{y}_{j}-\mathbf{y}_{t}.
\end{equation}
\end{linenomath*}

\subsection{Experiments on Other Datasets}
We also perform MI attacks on the digit recognition task, the object classification task, and the disease prediction task, using the MNIST~\cite{lecun1998gradient}, CIFAR-10~\cite{krizhevsky2009learning}, and ChestX-Ray~\cite{wang2017chestx} datasets for experiments, respectively.

\paragraph{Experimental Details.} 
For MNIST and CIFAR-10, we use the images in the training data with labels 0, 1, 2, 3, and 4 as the private data, containing 30,596 and 25,000 images, respectively. The rest images with labels 5, 6, 7, 8, and 9 are used as the public data, containing 29,404 and 25,000 images, respectively. We adopt ResNet-18 trained on the private data as the target model.  As for the evaluation model, we train VGG16 on the original training data instead of the private data to better distinguish deceptive reconstructed images. For ChestX-Ray, we use 14 classes of images with diseases as the private data to train the target model ResNet-18. Then, we randomly select 20,000 images from the class with the label “NoFinding” (i.e., no disease) as the public data to train the evaluation model ResNet-34. Furthermore, we also use a different COVID19 dataset~\cite{cohen2020covid} with 21,165 images as the public data. All images from above datasets are resized to $64 \times 64$. $n$ for the top-\emph{n} selection strategy is set to 4,000 for MNIST and CIFAR-10, and 1,000 for ChestX-Ray. The learning rates in stage-2 are set to $0.1$ and $0.001$, respectively, for 300 iterations. We reconstruct 100 images per class for MNIST and CIFAR-10, and 10 images per class for Chest-X-ray.

\begin{figure}
\hspace{-4mm}
\includegraphics[width=1.0 \linewidth]{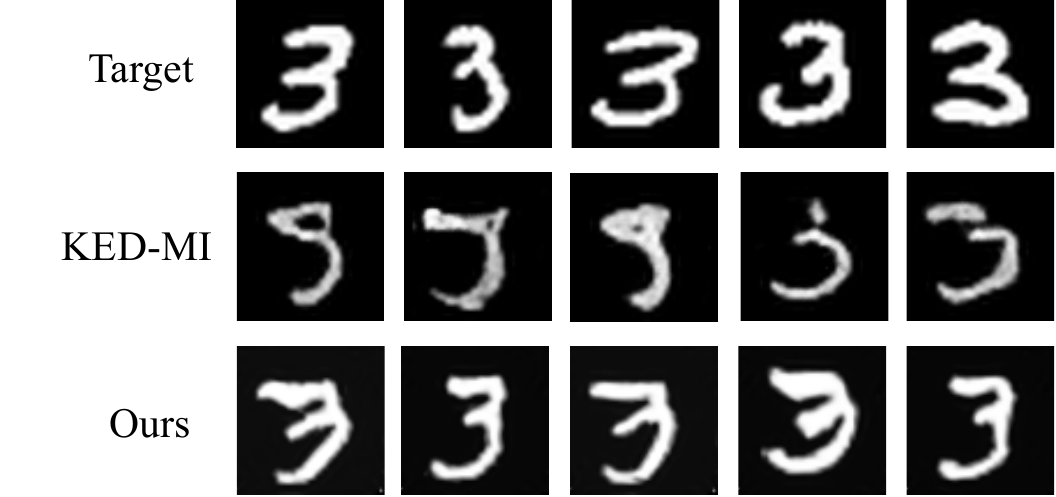}
\caption{The digit “3” samples of MNIST reconstructed by the baseline and our method.}
\label{fig:MNIST_images}
\end{figure}

\paragraph{Experimental Results.} 

\begin{figure}
\centering
\includegraphics[width=1.0 \linewidth]{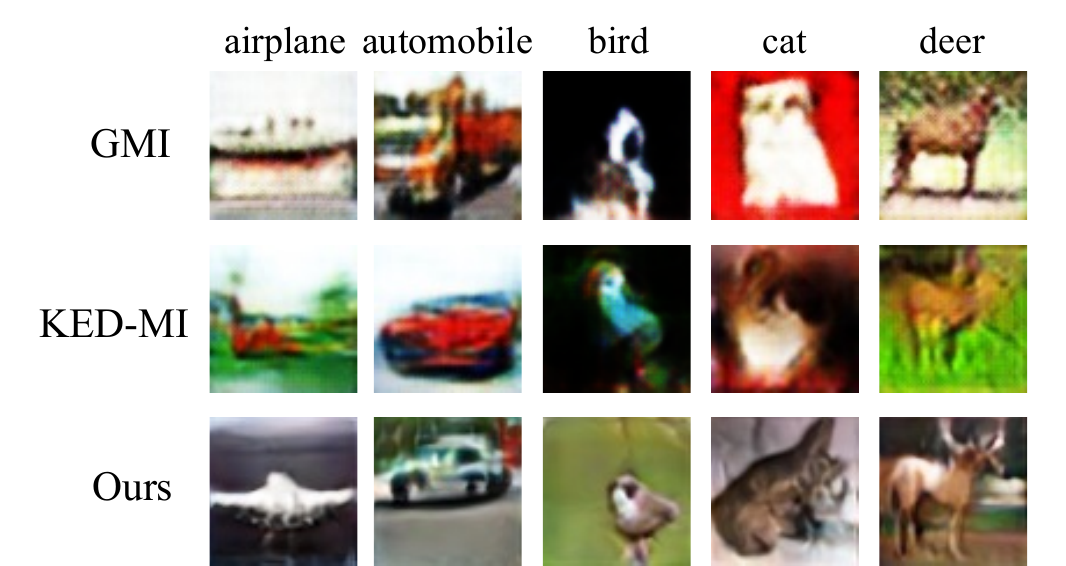}
\caption{CIFAR-10 samples reconstructed by the baselines and our method.}
\label{fig:CIFAR_images}
\end{figure}

As shown in~\Tref{tab:MNIST_CIFAR}, our method comprehensively outperforms the baselines by a large margin on MINST and CIFAR-10 datasets. When attacking the digit recognition model trained on MNIST, GMI performs very poorly. Although KED-MI can slightly improve the attack accuracy, it cannot reconstruct the private images of the digit “3” well, as shown in \Fref{fig:MNIST_images}. In contrast, our method reduces interference between different classes, thus allowing private images to be reconstructed with good visual quality on each class. 
Similar visual comparisons on CIFAR-10 are also shown in~\Fref{fig:CIFAR_images}. Our method not only significantly outperforms GMI and KED-MI in terms of image realism, but also recovers more accurate semantic information of private classes. It is difficult for GMI to reconstruct accurate private images, while KED-MI tends to reconstruct deceptive images that can be classified by the target model but have few human-recognizable features. 

\begin{figure*}
\centering

\includegraphics[width=0.75 \linewidth]{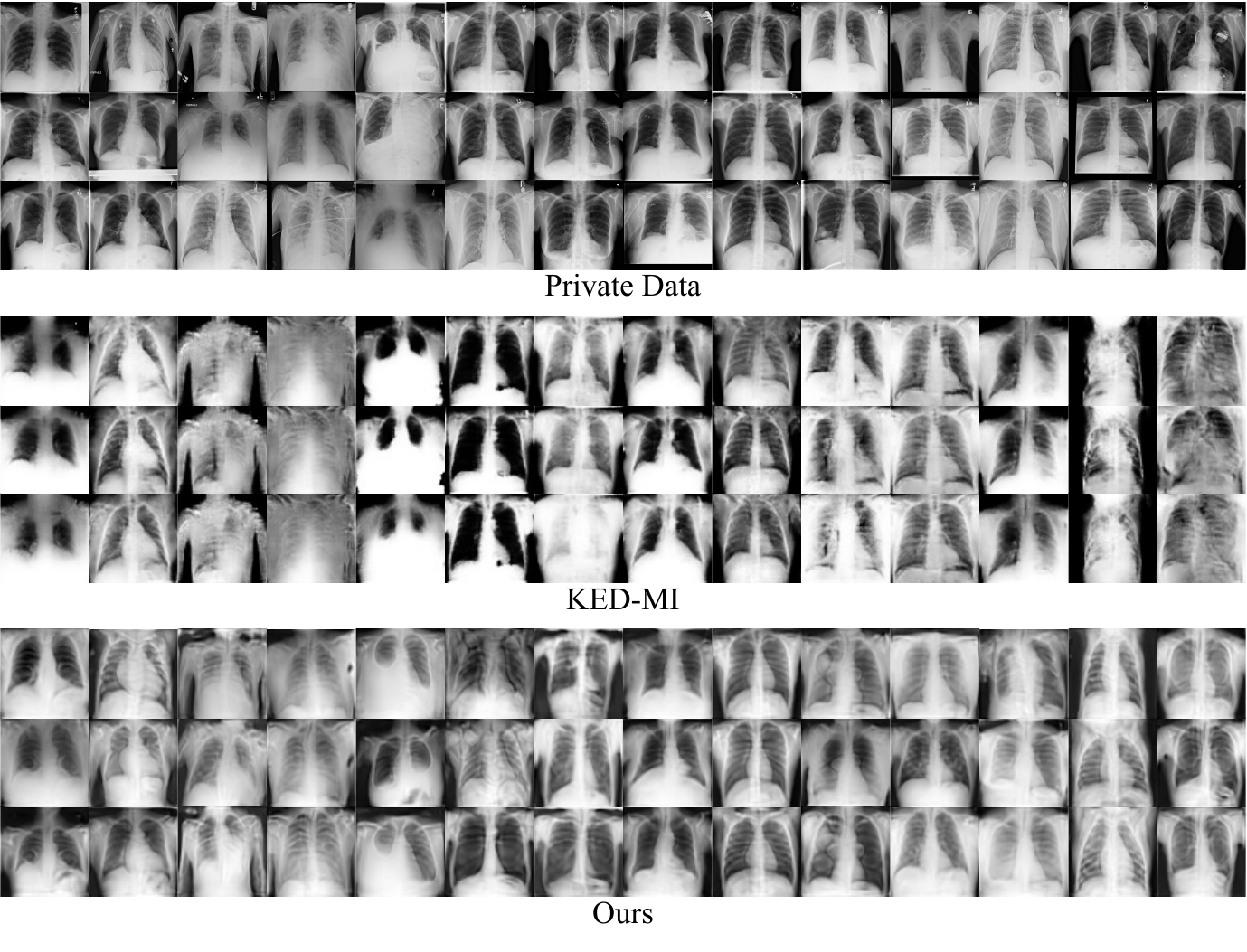}
\caption{ChestX-Ray samples reconstructed by the baseline (KED-MI) and our method when using COVID19 as the public dataset. Each column represents one category of the 14 diseases.}
\label{fig:ChestX_images}
\end{figure*}

The results of ChestX-Ray are shown in~\Tref{tab:ChestXRay}. Compared to KED-MI, our method improves the attack accuracy by an average of 12\% under the two different public datasets. Meanwhile, the FID is even reduced from 109.41 to 47.51 when the public data is COVID19. We also provide the visual comparison of the private data reconstructed by KED-MI and our method, as shown in~\Fref{fig:ChestX_images}.

\begin{table}[!ht] \LARGE
		\begin{center}
			\scalebox{0.65}{ 
			\begin{tabular}{ccccc}
                \hline
                                                  &                 & \textbf{Attack Acc $\uparrow$} & \textbf{KNN Dist $\downarrow$} & \textbf{FID $\downarrow$} \\ \hline
                \multirow{3}{*}{\textbf{MNIST}}   & \textbf{GMI}    & .03$\pm$.0054             & 171.18              & 137.53         \\
                                                  & \textbf{KED-MI} & .29$\pm$.0228             & 78.67               & 116.47         \\
                                                  & \textbf{Ours}   & \textbf{.60$\pm$.0465}    & \textbf{38.33}      & \textbf{77.68} \\ \hline
                \multirow{3}{*}{\textbf{CIFAR10}} & \textbf{GMI}    & .11$\pm$.0173             & 38.50               & 198.18         \\
                                                  & \textbf{KED-MI} & .50$\pm$.0392             & 15.49               & 141.18         \\
                                                  & \textbf{Ours}   & \textbf{.76$\pm$.0318}    & \textbf{5.11}       & \textbf{83.43} \\ \hline
            \end{tabular}}
		\end{center}
		\caption{Attack performance comparison on MNIST and CIFAR10.}
		\label{tab:MNIST_CIFAR}
\end{table}	

\begin{table}[!ht] \LARGE
\centering
\scalebox{0.54}{
\begin{tabular}{ccccc}
                \hline
                \multirow{2}{*}{}     & \multicolumn{2}{c}{\textbf{NoFinding $\rightarrow$ ChestX-Ray}} & \multicolumn{2}{c}{\textbf{COVID19 $\rightarrow$ ChestX-Ray}} \\
                                              & \textbf{KED-MI}       & \textbf{Ours}            & \textbf{KED-MI}        & \textbf{Ours}            \\ \hline
                \textbf{Attack Acc $\uparrow$} & .71$\pm$.0034             & \textbf{.82$\pm$.0048}       & .68$\pm$.0026              & \textbf{.80$\pm$.0020}       \\
                \textbf{KNN Dist $\downarrow$}   & 97.26                 & \textbf{82.71}           & 113.18                 & \textbf{84.56}           \\
                \textbf{FID $\downarrow$}        & 97.35                 & \textbf{64.33}           & 109.41                 & \textbf{47.51}           \\ \hline
\end{tabular}}
\caption{Attack performance comparison with KED-MI on ChestX-Ray.}
\label{tab:ChestXRay}
\end{table}

\paragraph{Analysis on the Failure of KED-MI.} 
As mentioned in the main manuscript, the KED-MI adopts the semi-supervised GAN framework and uses the feature-matching loss to train the generator. However, as demonstrated in~\cite{salimans2016improved}, the feature-matching loss works better if the goal is to obtain a strong classifier (i.e., the discriminator) using the semi-supervised GAN, otherwise it reduces the visual quality of the generated images. Moreover, Dai \etal~\cite{dai2017good} theoretically analyzes that there is a trade-off between the quality of the discriminator and the generator. 
Obviously, this framework is not suitable for generative MI attacks, since what we need in stage-2 is a good and accurate generator, contrary to the goal of semi-supervised GAN.

Rather than searching the latent vector of each reconstructed image as GMI and our method, KED-MI proposed to search for a distribution $\mathcal{N}(\mu, \sigma ^ {2})$ with two learnable parameters $\mu$ and $\sigma ^ {2}$ for each target class, and the latent vectors can be obtained by randomly sampling from the learned distribution. However, it is very difficult to learn an accurate class distribution by $\mathcal{N}(\mu, \sigma ^ {2})$ when the inner-class distribution of the private dataset varies greatly (e.g., CIFAR-10). As a result, latent vectors sampled from such a distribution are far from the manifold of natural images, although the produced images may be close to private classes in feature space. Whereas we use a more powerful conditional GAN to model the distribution of each private class, making it possible to handle more diverse datasets. In addition, we also perform random transformations on the generated images during the search process to further filter deceptive or adversarial samples.

\subsection{Empirical Comparisons of Different $\mathcal{L}_{inv}$}
We note that a concurrent work~\cite{struppek2022ppa} also has a similar observation of the gradient vanishing problem when using cross-entropy loss in MI attacks. Inspired by~\cite{li2020towards}, they adopt a more complex poincaré loss as follows:
\begin{linenomath*}
\begin{equation}
    \begin{aligned} 
    \mathcal{L}_{\text {Poincaré }} &=d(\boldsymbol{u}, \boldsymbol{v}) \\ &=\operatorname{arcosh}\left(1+\frac{2\|\boldsymbol{u}-\boldsymbol{v}\|_{2}^{2}}{\left(1-\|\boldsymbol{u}\|_{2}^{2}\right)\left(1-\|\boldsymbol{v}\|_{2}^{2}\right)}\right), \\
    &\boldsymbol{u}=\frac{\boldsymbol{u}}{\left\| \boldsymbol{u}\right\|_{1}}, \boldsymbol{v}=\max \{\boldsymbol{v}-\xi, 0\},
    \end{aligned}\label{eq:poincare_loss}
\end{equation}
\end{linenomath*}
where $\boldsymbol{u}$ is the logits normalized by the $\ell_{1}$ distance, and $\boldsymbol{v}$ is the one-hot encoded target vector with respect to the target class. ${\xi}=10^{-5}$ is a small constant to ensure numerical stability. 

In the main manuscript, we theoretically compare the cross-entropy loss and the proposed max-margin loss. Here, we further empirically compare them with the additional poincaré loss mentioned in~\Eref{eq:poincare_loss}. Specifically, we take these three losses as $\mathcal{L}_{inv}$ respectively in the iterative process of stage-2, and plot the trend curves of gradient values, loss values, and target logit values. Since the gradient values of different losses have a gap, we rescale the gradient values by dividing $\left\|g_{0}\right\|_{1}$, where $g_{0}$ is the gradient of the first iteration. The $\ell_{1}$ norm of the gradient $\left\|g_{i}\right\|_{1}$ represents its gradient value. Similarly, the loss values are also rescaled by dividing its value of the first iteration. We use FFHQ and CelebA as the public and private dataset, respectively. Then, we use VGG16 as the target model to attack the first 100 classes of CelebA and take their average results for plotting. 

\begin{figure*}
\centering 
\includegraphics[width=1.0 \linewidth]{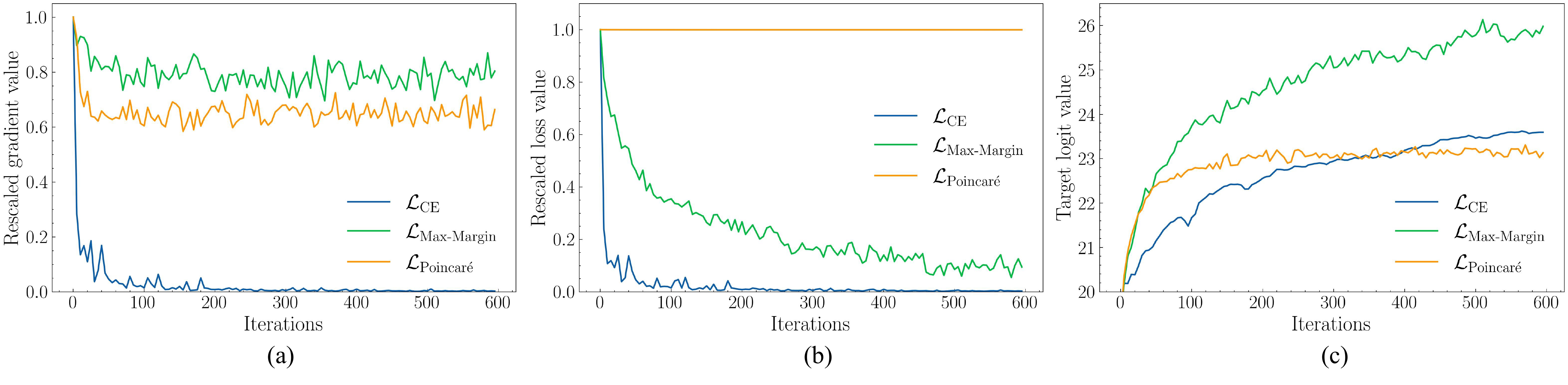}
\caption{(a)-(c) are the trend curves of the rescaled gradient values, the rescaled loss values, and the target logit values for different losses over the iterations, respectively.}
\label{fig:trend_curves}
\end{figure*}

\begin{figure}
\centering
\includegraphics[width=0.7 \linewidth]{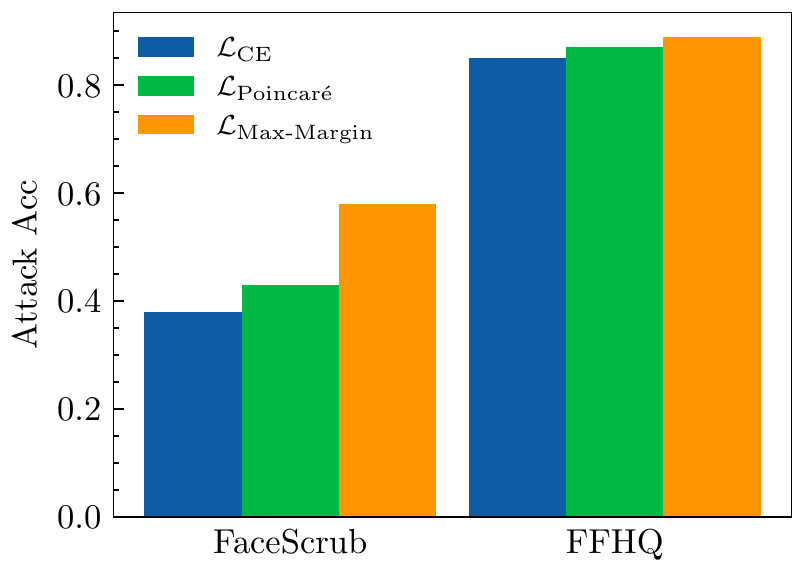}
\caption{Comparison of attack performance when using different losses in the image reconstruction stage.}
\label{fig:compare_with_poincare}
\end{figure}

From~\Fref{fig:trend_curves} (a), we can see that the gradient of the cross-entropy loss quickly decreases to 0 as the number of iterations increases, indicating the existence of the gradient vanishing problem. Meanwhile, the max-margin loss maintains a more stable gradient magnitude than the poincaré loss over the iterations, i.e., the degradation is more slight. In~\Fref{fig:trend_curves} (b), the value of the cross-entropy loss also quickly decreases to 0 due to the gradient vanishing, while the value of poincaré loss keeps almost unchanged, which may make it difficult to judge the optimization situation. In contrast, the max-margin loss can be minimized continuously over the iterations with relatively stable gradients. As shown in \Fref{fig:trend_curves} (c), the max-margin loss can make the logit value of the target class steadily increase over the iterations to reach a higher value, while the cross-entropy loss and poincaré loss gradually tend to a relatively constant value in the later iterations. In addition, ~\Fref{fig:compare_with_poincare} shows the attack performance when using these three losses separately in stage-2 of our method, demonstrating that the max-margin loss outperforms the other two losses, especially when there is a larger distributional shift between public and private data.

\subsection{More Discussions on the Concurrent Work}

Apart from the loss function, the other differences between our method and the concurrent work~\cite{struppek2022ppa} are three-fold. First, the motivations are different: we propose to design a more powerful general MI attack framework, while they aim to omit the training process of stage-1 to achieve plug-and-play. Second, based on the different goals, the technical contributions are different: we propose a top-\emph{n} selection strategy and aim to train a class-independent generator to search private images more accurately, while they focus on extending the reconstruction process in stage-2 to reduce the dependence on the trained generator. Third, the application scenarios are different: what we propose is a general attack method that can be applied to tasks trained on various sensitive data (e.g, disease diagnosis), while their method relies on the task of the publicly available pre-trained GAN and can not applicable for some unique tasks. Overall, our method differs from ~\cite{struppek2022ppa} in terms of motivation, technical contributions, and application scenarios.

\end{document}